\documentclass{article}

\PassOptionsToPackage{numbers, compress}{natbib}
 \usepackage[preprint]{neurips_2026}

\usepackage[utf8]{inputenc} 
\usepackage[T1]{fontenc}    
\usepackage{hyperref}       
\usepackage{url}            
\usepackage{booktabs}       
\usepackage{amsfonts}       
\usepackage{bm}             
\usepackage{nicefrac}       
\usepackage{microtype}      
\usepackage{xcolor}         
\usepackage{amsmath}
\usepackage{amssymb}
\usepackage{adjustbox}
\usepackage{wrapfig}
\usepackage{multirow}
\usepackage[table]{xcolor}

\DeclareMathOperator*{\argmin}{arg\,min}

\usepackage[font=small,labelfont=bf]{caption}

\title{Knowledge Transfer Scaling Laws for 3D Medical Imaging}

\author{%
  Ho Hin Lee\thanks{Corresponding author: ho.hin.lee@vanderbilt.edu} \\
  Vanderbilt University\\
  \And
  Dongna Du \\
  Zhejiang University\\
  \And
  Chu Wang \\
  McGill University\\
  \And
  Yuankai Huo \\
  Vanderbilt University \\
  \And
  Shi Gu \\
  Zhejiang University\\
  \And
  James C. Gee \\
  University of Pennsylvania\\
  \And
  Yifan Wu\\
  University of Pennsylvania\\
}

\begin{document}

\maketitle

\begin{abstract}
Vision foundation models are increasingly moving beyond 2D to volumetric domains such as 3D medical imaging, where unified pretraining across different imaging modalities (i.e. CT, MRI, and PET) could provide foundational models for diverse clinical tasks. However, training such models requires mixing heterogeneous imaging domains, and current mixture strategies remain largely heuristic. In this work, we observe that different medical imaging domains scale at variable rates during pretraining, and knowledge transfer between domains is strongly asymmetric: training on one domain can substantially improve another, but the reverse may be much weaker. Interestingly, both MAE reconstruction loss and cross-domain transfer follow predictable power-law trends with domain-specific behaviors. Motivated by these findings, we formulate data allocation as a \textbf{scaling-law optimization problem}. The derived allocations reveal an interpretable \textbf{hub-and-island} structure: highly transferable domains emerge as \emph{hubs} that benefit many others and deserve strategic allocation, while isolated domains act as \emph{islands} requiring direct investment. Empirically, transfer-aware allocation outperforms data-proportional sampling by up to $58\%$ and generalizes well to unseen budgets with $r{=}0.989$. Downstream validation on disease classification and organ/lesion segmentation further confirms that the derived transfer-aware mixtures provide stronger pretrained representations for clinical 3D medical imaging tasks.
\end{abstract}

\begin{figure*}[h]
\centering
\includegraphics[width=\textwidth]{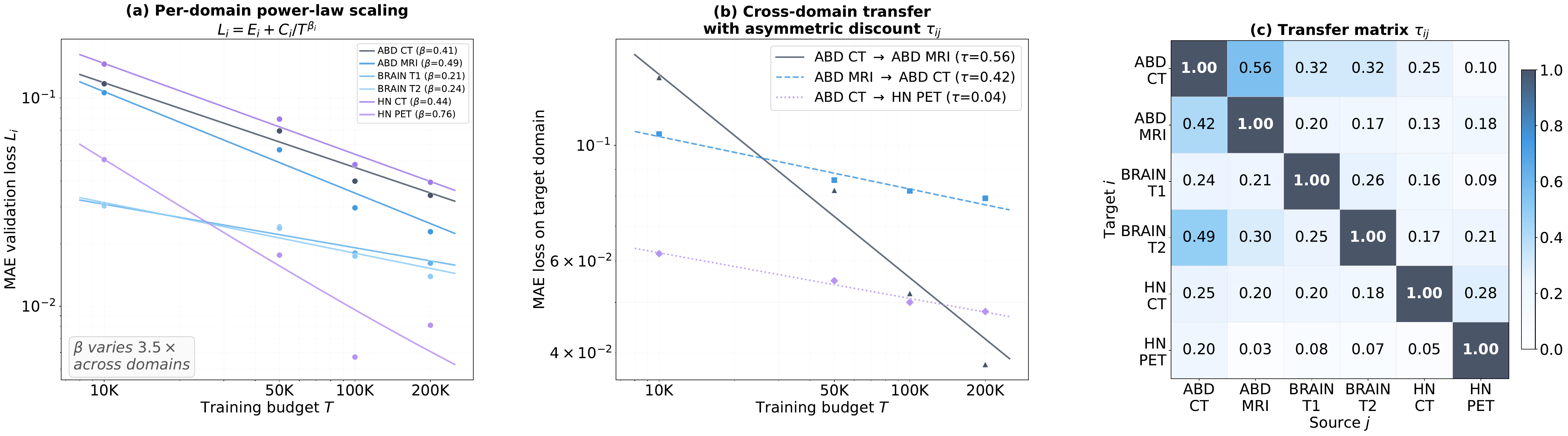}
\caption{
\textbf{Observations motivating the transfer-aware scaling law.}
\textbf{(a)} Per-domain MAE loss follows 
$L_i = E_i + C_i / T^{\beta_i}$ with domain-specific exponents 
varying $3.5\times$ ($\beta = 0.21$ for BRAIN T1 to 
$\beta = 0.76$ for HN PET), indicating that domains benefit 
from additional data at fundamentally different rates.
\textbf{(b)} Cross-domain transfer obeys the same power-law 
form with a directional discount $\tau_{ij}$: ABD CT transfers 
strongly to ABD MRI ($\tau{=}0.56$) but weakly to HN PET 
($\tau{=}0.04$), and the reverse ABD MRI $\rightarrow$ ABD CT 
($\tau{=}0.42$, dashed) is weaker than the forward 
direction.
\textbf{(c)} Full $6{\times}6$ transfer matrix $\tau_{ij}$ 
(from source $j$ to target $i$) estimated from pure-domain 
proxy runs. The matrix is asymmetric ($\tau_{ij} \neq \tau_{ji}$) 
and reveals hub-and-island structure: ABD CT (hub) transfers broadly 
to most domains, while HN PET (island) is largely isolated.
}
\label{fig:observation}
\end{figure*}

\section{Introduction}

Clinical imaging spans CT, MRI, and PET, each with distinct acquisition physics, resolution, and clinical signal, yet the fundamental skills required to interpret them (volumetric reasoning, structural context, cross-anatomy knowledge) are shared. A \textbf{unified} 3D medical foundation model that learns reusable representations across modalities and anatomies could support diverse clinical tasks, from organ/lesion segmentation to abnormality discovery and disease monitoring from a single pretrained initialization~\cite{rui2025multi,xu2025generalizable,tang2022self,wald2025openmind}. However, training such a unified model requires large-scale heterogeneous volumetric datasets with diversity, but while architectures and self-supervised objectives have received substantial attention, a more fundamental question remains open: \textbf{how much of each imaging domain should be used under a fixed training token budget?}

In natural language processing, scaling laws~\cite{kaplan2020scaling,hoffmann2022training} have enabled principled data mixture optimization by modeling how performance scales with compute and data~\cite{ye2024data,ilse2025data,shukor2025scaling,xie2023doremi}. However, these formulations assume relatively homogeneous data sources or shared scaling behavior, and have seen limited adoption in vision settings, particularly in 3D medical imaging. In practice, mixtures are still chosen by heuristics (i.e.\ uniform or data-proportional sampling) that treat all imaging domains as interchangeable. However, as shown in Figure~\ref{fig:observation}(a), we observe that each imaging domain exhibit substantially different scaling behavior and the corresponding self-supervised masked autoencoding (MAE) reconstruction loss follows a predictable power law $L \approx E + C / T^{\beta}$, where the scaling exponent~$\beta$ controls how rapidly a domain benefits from additional data, but varies by $3.5\times$ across domains. Furthermore, Figure~\ref{fig:observation}(b) demonstrates that cross-domain transfer also follows the power law discounted by an asymmetric coefficient~$\tau_{ij}$. Such two regularities \textbf{heterogeneous scaling} and \textbf{directional transfer}, provide the analytical foundation to optimize data mixtures in closed form.

We therefore formulate 3D medical data mixture design as a \textbf{scaling-law optimization problem}~\cite{kaplan2020scaling,hoffmann2022training,shukor2025scaling,ilse2025data}. Rather than directly fitting downstream segmentation performance, we use masked autoencoding (MAE) as a controlled proxy objective that can be applied consistently across heterogeneous 3D imaging domains~\cite{he2022masked}. While reconstruction is not the end goal, the quality of learned representations, as measured by reconstruction loss, directly determines how well the pretrained model supports downstream clinical tasks. We propose a \textbf{transfer-aware scaling law} with two ingredients: \emph{(i)} domain-specific scaling exponents~$\beta_i$, which capture how efficiently each domain benefits from additional training budget; and \emph{(ii)} an asymmetric transfer matrix~$\tau_{ij}$, which measures how effectively training on source domain~$j$ contributes to target domain~$i$. Together, these components define a closed-form surrogate for predicting domain-wise MAE loss under arbitrary mixtures. Optimizing this surrogate yields a compute-dependent allocation~$\bm{h}^*(T)$ without without exhaustive and compute hungry data mixture ablation experiments.

The resulting allocations reveal interpretable \textbf{hub-and-island} structure: domains with broad outgoing transfer act as \emph{hubs}, where compute allocated benefits multiple domains simultaneously, while domains with weak incoming transfer act as \emph{islands} that require direct investment, as knowledge from hubs cannot be transferred to them effectively. Such deterministic knowledge base structure shows that dataset-size weighting alone is insufficient, where data recipe design must account for both domain-specific scaling and cross-domain transfer. Empirically, across six heterogeneous 3D imaging domains spanning CT, MRI, and PET, the estimated transfer matrix reveals strong asymmetry and the derived optimal
allocations differ substantially from uniform and data-proportional heuristics. By scaling up the token budget for pretraining, transfer-aware allocation achieves the lowest MAE loss, outperforming data-proportional allocation by up to $58\%$, and extrapolates to unseen budgets with strong per-domain agreement ($r=0.989$). 
We further validate the derived optimal allocation, whose transfer matrices are estimated from principled MAE proxy experiments, on downstream clinical tasks such as abnormal findings classification and lesion segmentation. Our results demonstrate that the proposed transfer-aware data recipe provides a strong prior for supervised 3D medical imaging tasks. In summary, our main contributions are:
\begin{itemize}
    \item We formulate cross-domain data allocation in 3D medical imaging as a scaling-law optimization problem, using MAE reconstruction loss as a controlled proxy objective across six heterogeneous imaging domains.
    \item We propose a transfer-aware scaling law that jointly models heterogeneous scaling exponents and an asymmetric transfer matrix, yielding a closed-form surrogate for compute-dependent mixture optimization from small-scale proxy runs.
    \item We show that the learned allocations reveal interpretable hub-and-island structure, outperform standard heuristic mixtures across budgets, extrapolate from small proxy runs to unseen budgets, and provide a useful prior for downstream tasks.
\end{itemize}

\section{Related Work}
\label{sec:related_work}

\textbf{3D medical foundation models and self-supervised pretraining.}
Self-supervised pretraining has become a key strategy for learning transferable representations from unlabeled 3D medical images, where dense annotations are costly. Early work explored restoration, contrastive, and masked prediction objectives for volumetric data~\cite{zhou2021models,taleb20203d,tang2022self}, while recent foundation-style models scale pretraining to larger and more heterogeneous corpora, including CT volumes, multi-parametric MRI, and diverse multi-domain medical datasets~\cite{wang2023mis,rui2025multi,xu2025generalizable,wald2025openmind}. These studies show that heterogeneous 3D data can support reusable representations across tasks and domains. However, they primarily focus on architectures \cite{isensee2021nnu, hatamizadeh2022unetr, hatamizadeh2022swin,lee20223d, yu2023unest, lee2023scaling,lee2025rep3d}, objectives \cite{wald2025openmind, krishnan2022self, xie2022unimiss, ye2024continual}, and evaluation protocols \cite{wang2024fedmeki,jin2024fairmedfm,wang2023real,bai2023benchmarking,bassi2024touchstone}, while the data mixture is typically fixed by corpus construction or heuristic sampling rather than optimized. \\
\textbf{Scaling laws and data-mixture optimization.}
Scaling laws provide a quantitative framework for predicting how loss changes with model size, data, and compute~\cite{kaplan2020scaling,hoffmann2022training}. Recent data-centric extensions further show that training mixtures can be estimated from small-scale proxy runs instead of exhaustively swept, including data mixing laws, RegMix, BiMix, and optimal mixture scaling formulations~\cite{ye2024data,liu2024regmix,ge2024bimix,shukor2025scaling}. Related work on multilingual and general-purpose pretraining also studies proportional, temperature-based, or learned reweighting strategies for imbalanced corpora~\cite{arivazhagan2019massively,xue2021mt5,chung2023unimax,xie2023doremi}. However, these approaches are largely developed for language, general vision, or native multimodal pretraining, and generally do not model the asymmetric source-to-target transfer that arises across heterogeneous 3D medical domains.\\
\textbf{Cross-domain transfer in medical imaging.}
Medical imaging domains differ substantially in acquisition physics, spatial resolution, anatomical coverage, and clinical signal. Prior multi-domain and multi-modal work has shown that joint training can improve transfer and generalization, including Med3D, Hermes, and recent 3D foundation models~\cite{chen2019med3d,gao2024training,rui2025multi,xu2025generalizable}. More broadly, transfer effectiveness can be predictable and directional~\cite{hernandez2021scaling}. Nevertheless, most medical imaging studies evaluate transfer under a fixed mixture, rather than parameterizing transfer as a quantity for allocation. In contrast, we estimate an asymmetric transfer matrix and use it to optimize compute-aware data allocation across 3D medical domains.

\section{Method}
\label{sec:method}

We formulate heterogeneous 3D medical data mixture design as a transfer-aware scaling-law optimization problem.
Given a fixed self-supervised pretraining budget, our goal is to predict the masked autoencoding (MAE) validation loss induced by a candidate data mixture and choose the mixture that minimizes the expected loss across target domains.
Our method is built on two observations.
First, different 3D medical domains exhibit different scaling behavior: some domains continue to benefit from additional data, while others saturate more quickly.
Second, cross-domain transfer is directional: training on source domain $j$ may benefit target domain $i$ differently from how training on $i$ benefits $j$.
We capture these effects with domain-specific scaling exponents and an asymmetric transfer matrix.

\subsection{Problem setup}
\label{sec:problem_setup}

We consider self-supervised pretraining on $K$ heterogeneous 3D medical imaging domains
$\mathcal{D}_1,\dots,\mathcal{D}_K$. Each domain corresponds to a modality-anatomy pair, such as abdominal CT, abdominal MRI, brain MRI, or head-and-neck PET (HN-PET). Given a domain weight vector
$\bm{h}=(h_1,\dots,h_K)$, we define the mixture distribution
\begin{equation}
    \mathrm{mix}(\bm{h})
    =
    \sum_{j=1}^K h_j \mathcal{D}_j,
    \qquad
    \bm{h}\in\Delta_\epsilon^{K-1},
\end{equation}
where
\begin{equation}
    \Delta_\epsilon^{K-1}
    =
    \left\{
    \bm{h}\in\mathbb{R}^K:
    h_j\ge\epsilon,\;
    \sum_{j=1}^K h_j=1
    \right\}.
\end{equation}
Thus, domain $j$ is sampled with probability $h_j$.
The lower bound $\epsilon$ prevents degenerate zero-allocation mixtures and ensures that every domain remains represented during mixture optimization.

Let $T$ denote the total pretraining budget, measured in volume draws.
Training with mixture $\bm{h}$ allocates $h_jT$ expected samples to domain $j$.
We train a 3D masked autoencoder with parameters $\theta$ using samples from $\mathrm{mix}(\bm{h})$.
Let $\ell(x,\theta)$ denote the reconstruction loss for a 3D volume $x$.
For a target domain $\mathcal{D}_i$, the target-domain MAE loss is
\begin{equation}
    L_i(\theta)
    =
    \mathbb{E}_{x\sim \mathcal{D}_i}
    \left[
    \ell(x,\theta)
    \right].
\end{equation}

After training on mixture $\bm{h}$ for budget $T$, the optimization procedure returns parameters
$\theta^*(\bm{h},T)$.
The loss induced by this training run on target domain $i$ is
\begin{equation}
    L_i(\bm{h},T)
    =
    L_i\left(\theta^*(\bm{h},T)\right).
\end{equation}
Our goal is to predict $L_i(\bm{h},T)$ for each target domain without exhaustively training models under many candidate mixtures. We therefore learn a surrogate $\widehat{L}_i(\bm{h},T)$ for each target domain and choose the mixture that minimizes the average predicted loss:
\begin{equation}
    \bm{h}^*(T)
    \in
    \argmin_{\bm{h}\in\Delta_\epsilon^{K-1}}
    \frac{1}{K}
    \sum_{i=1}^K
    \widehat{L}_i(\bm{h},T).
    \label{eq:optimal_mixture_main}
\end{equation}
This objective treats all target domains equally.
A weighted version is introduced in Section~\ref{sec:fitting_and_optimization}.

\subsection{Transfer-aware scaling law}
\label{sec:transfer_aware_scaling_law}

We now derive the predictive law used in Eq.~\eqref{eq:optimal_mixture_main}.
The derivation has three components:
domain-specific self-scaling, directed cross-domain transfer, and a mixture-level effective budget.

\paragraph{Domain-specific self-scaling.}
We first model how each domain scales when trained in isolation.
When the model is trained only on samples from $\mathcal{D}_i$, we model the validation MAE loss on the same domain using a power law:
\begin{equation}
    L_i(T)
    =
    E_i
    +
    \frac{C_i}{T^{\beta_i}},
    \label{eq:self_domain_scaling}
\end{equation}
where $E_i\ge0$ is the asymptotic loss floor, $C_i>0$ controls the reducible loss, and $\beta_i>0$ is the scaling exponent for target domain $i$. The exponent $\beta_i$ controls the marginal benefit of additional effective data. A larger $\beta_i$ indicates that the domain continues to improve more rapidly with additional pretraining samples, whereas a smaller $\beta_i$ indicates faster saturation.
Unlike mixture laws that assume a shared exponent across domains, we allow $\beta_i$ to vary with the target domain. This is important for 3D medical imaging because the domains differ not only in dataset size, but also in modality, anatomical coverage, spatial resolution, and signal characteristics. The self-scaling law therefore captures the intrinsic data-scaling behavior of each target domain before considering transfer.

\paragraph{Directed cross-domain transfer.}
Self-domain scaling alone only accounts for samples drawn from the same domain as the target.
However, pretraining on one 3D medical domain can improve representations for another domain.
We model this effect through a directed transfer matrix
$\bm{\tau}\in[0,1]^{K\times K}$.
The entry $\tau_{ij}$ measures how effectively training on source domain $\mathcal{D}_j$ contributes to the target loss on $\mathcal{D}_i$. We set $\tau_{ii}=1$ by definition. For $i\ne j$, $\tau_{ij}$ converts source-domain samples into target-domain effective samples. When the model is trained only on source domain $\mathcal{D}_j$, we predict the loss on target domain $\mathcal{D}_i$ by reusing the target-domain scaling parameters and discounting the budget:
\begin{equation}
    L_{i\leftarrow j}(T)
    =
    E_i
    +
    \frac{C_i}{(\tau_{ij}T)^{\beta_i}}.
    \label{eq:directed_transfer_scaling}
\end{equation}
This gives $\tau_{ij}$ a direct interpretation:
$T$ samples from source domain $j$ are equivalent, for target domain $i$, to $\tau_{ij}T$ samples from the target domain itself.
A value close to one indicates strong transfer, while a value near zero indicates weak transfer. As the matrix is directed, so we do not require $\tau_{ij}=\tau_{ji}$. Such directionality is central to our formulation.
For example, a source domain with strong anatomical structure may provide useful representations for several targets, while a specialized target domain may not provide the same benefit in reverse.
By estimating $\tau_{ij}$ for each ordered pair, the model can distinguish transferable sources from domains that require direct allocation.

\paragraph{Transfer-aware mixture law.}
For a general mixture $\bm{h}$, source domain $j$ receives $h_jT$ samples. Its contribution to target domain $i$ is discounted by $\tau_{ij}$ (the transfer coefficient \emph{from} source~$j$ \emph{to} target~$i$), yielding $\tau_{ij}h_jT$ effective target-domain samples.
The total effective budget reaching target domain $i$ is therefore
\begin{equation}
    T_{\mathrm{eff},i}(\bm{h},T)
    =
    \sum_{j=1}^K
    \tau_{ij}h_jT.
    \label{eq:effective_budget_main}
\end{equation}
Substituting this effective budget into the self-domain scaling law in Eq.~\eqref{eq:self_domain_scaling} gives our transfer-aware scaling law:
\begin{equation}
    \widehat{L}_i(\bm{h},T)
    =
    E_i
    +
    \frac{C_i}
    {
    \left(
    \sum_{j=1}^K
    \tau_{ij}h_jT
    \right)^{\beta_i}
    }.
    \label{eq:transfer_aware_law}
\end{equation}

Equation~\eqref{eq:transfer_aware_law} is the central surrogate used for mixture allocation.
It is closed-form, differentiable with respect to $\bm{h}$, and can be evaluated for arbitrary budgets once its parameters are estimated.
The law separates two effects that are entangled in direct mixture sweeps:
the target-domain scaling rate, controlled by $\beta_i$, and the source-to-target transfer structure, controlled by $\tau_{ij}$.
This separation allows us to ask whether a domain should receive budget because it is a high-value target, because it is a transferable source, or because it cannot be reached effectively through other sources.

\paragraph{Special cases.}
Our formulation recovers simpler data-mixture laws as special cases.
If cross-domain transfer is disabled by setting $\tau_{ij}=\delta_{ij}$, where $\delta_{ij}$ is the Kronecker delta 
($\delta_{ij}=1$ if $i=j$ and $0$ otherwise), then each target receives effective budget only from its own domain:
\begin{equation}
    \widehat{L}_i(\bm{h},T)
    =
    E_i
    +
    \frac{C_i}{(h_iT)^{\beta_i}}.
    \label{eq:transfer_naive_law}
\end{equation}
We refer to this as a transfer-naive mixture law.
If we further constrain all domains to share the same exponent, $\beta_i=\beta$, the model reduces to a shared-exponent mixture law.
Our formulation therefore generalizes these alternatives by combining domain-specific scaling and directed cross-domain transfer.

\subsection{Fitting and optimal mixture estimation}
\label{sec:fitting_and_optimization}

Once the functional form in Eq.~\eqref{eq:transfer_aware_law} is specified, we estimate its parameters from small-scale pure-domain pretraining runs.
We then optimize the fitted surrogate over the mixture simplex.
The fitting procedure has two stages.

\paragraph{Fitting self-domain scaling parameters.}
For each domain $i$, we train using only samples from 
$\mathcal{D}_i$ at several small-scale data budgets and record 
the validation MAE loss across random seeds. We fit 
$(E_i,C_i,\beta_i)$ by minimizing:
\begin{equation}
    \min_{E_i,C_i,\beta_i}
    \;\mathbb{E}_{d,\,e}
    \left[
    L_i^{(d,e)}
    -
    E_i
    -
    \frac{C_i}{d^{\beta_i}}
    \right]^2,
    \label{eq:self_fit_main}
\end{equation}
where the expectation averages uniformly over budgets $d$ and 
seeds $e$. We solve this bounded nonlinear least-squares problem 
using trust-region reflective 
optimization~\cite{branch1999subspace}, with constraints 
$E_i\ge0$, $C_i>0$, and $\beta_i\in[0.01,2.0]$.

\paragraph{Fitting directed transfer coefficients.}
For each directed pair $(i,j)$ with $i\ne j$, we train on source 
domain $\mathcal{D}_j$ and evaluate on target domain 
$\mathcal{D}_i$. Holding the target-domain scaling parameters 
$(E_i,C_i,\beta_i)$ fixed, we estimate the transfer coefficient 
$\tau_{ij}$ (\emph{from} source~$j$ \emph{to} target~$i$) by 
minimizing the expected squared residual over training budgets 
and random seeds:
\begin{equation}
    \hat{\tau}_{ij}
    =
    \argmin_{\tau\ge0}
    \;\mathbb{E}_{d,\,e}
    \left[
    L_{i\leftarrow j}^{(d,e)}
    -
    E_i
    -
    \frac{C_i}{(\tau d)^{\beta_i}}
    \right]^2,
    \label{eq:transfer_fit_main}
\end{equation}
where $d$ indexes the training budget and $e$ indexes random 
seeds at that budget. The expectation weights each budget level 
equally regardless of the number of seeds evaluated, ensuring 
the fit is unbiased across budget scales. This two-stage 
procedure separates intrinsic target-domain scaling from 
cross-domain transfer and avoids jointly fitting all $K^2+2K$ 
parameters. Each off-diagonal transfer coefficient is fit 
independently using the same bounded trust-region reflective 
solver.

\paragraph{Parameter count and estimation cost.}
The model requires $3K$ self-domain parameters
$\{E_i,C_i,\beta_i\}_{i=1}^K$
and $K(K-1)$ off-diagonal transfer coefficients
$\{\tau_{ij}\}_{i\ne j}$.
For $K$ domains, the fitting procedure uses $K$ self-domain configurations and $K(K-1)$ directed source-to-target configurations, for a total of $K^2$ pure-domain configurations.
These runs are performed at small scale and are used only to estimate the scaling-law surrogate.
Once fitted, the surrogate can be optimized for new budgets without additional mixture-training sweeps.

\paragraph{Estimating the optimal mixture.}
With all knowledge transfer scaling-law parameters $\tau_{ij}$'s estimated, we solve for the data mixture weights $\bm{h} = (h_1, \dots, h_K)$ that minimize the predicted loss across all target domains. Recall that $h_j$ specifies the fraction of the total training budget~$T$ allocated to domain~$j$, with $\sum_j h_j = 1$ and $h_j \ge \epsilon$. Each target domain~$i$ receives an effective budget computed by Eq.~\eqref{eq:effective_budget_main} with cross-domain transfer matrix $\tau_{ij}$, and the surrogate $\widehat{L}_i(\bm{h}, T)$ predicts the resulting MAE loss. The optimal mixture minimizes the weighted sum of predicted losses across targets:
\begin{equation}
    \bm{h}^*(T)
    \in
    \argmin_{\bm{h}\in\Delta_\epsilon^{K-1}}
    \sum_{i=1}^K
    w_i
    \widehat{L}_i(\bm{h},T),
    \label{eq:weighted_optimal_mixture}
\end{equation}
where $w_i \ge 0$ controls the relative importance of target domain~$i$ in reconstruction loss. Setting $w_i =1/K$ weights all domains equally, which we use throughout our experiments. Unequal weights allow practitioners to prioritize specific clinical domains if desired. As $\widehat{L}_i(\bm{h},T)$ is closed-form and differentiable in $\bm{h}$, we optimize Eq.~\eqref{eq:weighted_optimal_mixture} using gradient-based constrained optimization. We use SLSQP~\cite{kraft1988software}, as it natively enforces both the equality constraint $\sum_j h_j = 1$ and inequality constraints $h_j \ge \epsilon$ that standard unconstrained solvers (e.g., L-BFGS) cannot handle directly \cite{shukor2025scaling}. The resulting allocation is budget-dependent: as $T$ changes, the marginal value of assigning samples to each source domain shifts through both the target-domain exponent $\beta_i$ and the transfer matrix~$\bm{\tau}$, yielding a compute-dependent allocation rule rather than a single static mixture.

\begin{figure*}[t]
\centering
\includegraphics[width=\textwidth]{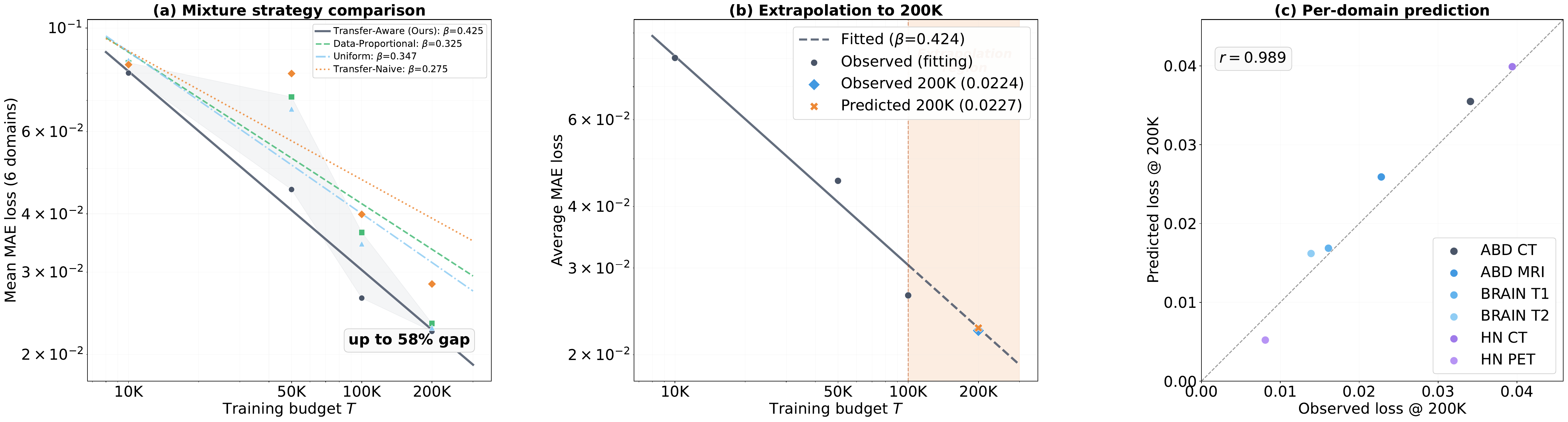}
\caption{
\textbf{Scaling law validation and extrapolation.}
\textbf{(a)} Transfer-Aware allocation achieves the steepest 
scaling ($\beta = 0.425$) and lowest MAE loss across all 
training budgets, outperforming Data-Proportional by up to 
$58\%$.
\textbf{(b)} The scaling law fitted on $\{10K, 50K, 100K\}$ 
extrapolates to the held-out $200K$ budget within $1.2\%$ error.
\textbf{(c)} Per-domain predicted vs.\ observed loss at 
$B = 200K$ shows strong agreement across all six domains 
($r = 0.989$).
}
\label{fig:main_results}
\end{figure*}

\begin{table*}[t]
\centering
\caption{
\textbf{Effective budget decomposition under Transfer-Aware allocation at $T{=}100K$.}
The effective budget reaching target domain $i$ is $T_{\mathrm{eff},i} = \sum_{j=1}^{K} \tau_{ij} \cdot h_j \cdot T$,
where each source domain $j$ contributes $\tau_{ij} \cdot h_j \cdot T$ effective samples.
Self-contributions ($\tau_{ii}{=}1$) are shown in \textbf{bold}.
ABD\_CT (hub, $h^*{=}23.7\%$) provides the largest cross-domain contribution to four targets.
HEAD\_PET (island, $h^*{=}56.3\%$) receives almost no incoming transfer and relies on its large direct allocation.
Floor domains ($h^*{=}5\%$) achieve $3.7$--$6.1\times$ amplification through incoming transfer.
}
\label{tab:teff_decomposition}
\vspace{4pt}
\resizebox{\textwidth}{!}{%
\begin{tabular}{l cccccc r c}
\toprule
& \multicolumn{6}{c}{\textbf{Source contribution} $\tau_{ij} \cdot h_j \cdot T$} & & \\
\cmidrule(lr){2-7}
\textbf{Target} $i$ & ABD-CT & ABD-MRI & BRAIN-T1 & BRAIN-T2 & HN-CT & HN-PET & $T_{\mathrm{eff},i}$ & Amplif. \\
& \small{($h{=}23.7\%$)} & \small{($h{=}5\%$)} & \small{($h{=}5\%$)} & \small{($h{=}5\%$)} & \small{($h{=}5\%$)} & \small{($h{=}56.3\%$)} & & \\
\midrule
ABD\_CT \small{(hub)}
  & \textbf{23,700} & 2,800 & 1,200 & 2,450 & 1,250 & 2,252
  & 33,652 & $1.4\times$ \\
ABD\_MRI
  & 11,139 & \textbf{5,000} & 800 & 1,600 & 700 & 1,689
  & 20,928 & $4.2\times$ \\
BRAIN\_T1
  & 7,584 & 1,000 & \textbf{5,000} & 1,300 & 1,950 & 3,378
  & 20,212 & $4.0\times$ \\
BRAIN\_T2
  & 8,295 & 1,300 & 1,100 & \textbf{5,000} & 950 & 1,689
  & 18,334 & $3.7\times$ \\
HEAD\_CT
  & 5,925 & 700 & 1,950 & 950 & \textbf{5,000} & 15,764
  & 30,289 & $6.1\times$ \\
HEAD\_PET \small{(island)}
  & 948 & 150 & 300 & 150 & 1,400 & \textbf{56,300}
  & 59,248 & $1.1\times$ \\
\bottomrule
\end{tabular}
}
\end{table*}

\begin{table}[t]
\centering
\caption{
Pretraining MAE reconstruction loss ($\downarrow$).
Left: average loss across six domains at different budgets.
Right: per-domain loss at $B=100K$.
Gray values denote relative increase over Transfer-Aware.
}
\label{tab:pretraining_mae}
\resizebox{\textwidth}{!}{
\begin{tabular}{l|cccc|cccccc|c}
\toprule
\multirow{2}{*}{\textbf{Datamix Strategies}}
& \multicolumn{4}{c|}{\textbf{Mean MAE across budgets}}
& \multicolumn{7}{c}{\textbf{Per-domain MAE at $B=100K$}} \\
\cmidrule(lr){2-5} \cmidrule(lr){6-12}
& $10K$ & $50K$ & $100K$ & $200K$
& ABD-CT & ABD-MRI & Brain-T1 & Brain-T2 & HN-CT & HN-PET & Mean \\
\midrule
Uniform
& 0.0853 {\scriptsize\textcolor{gray}{(+7\%)}}
& 0.0671 {\scriptsize\textcolor{gray}{(+49\%)}}
& 0.0345 {\scriptsize\textcolor{gray}{(+31\%)}}
& 0.0228 {\scriptsize\textcolor{gray}{(+2\%)}}
& 0.0499 & 0.0427 & 0.0202 & 0.0196 & 0.0641 & 0.0106 & 0.0345 \\
Data-Prop.
& 0.0843 {\scriptsize\textcolor{gray}{(+5\%)}}
& 0.0712 {\scriptsize\textcolor{gray}{(+58\%)}}
& 0.0365 {\scriptsize\textcolor{gray}{(+38\%)}}
& 0.0233 {\scriptsize\textcolor{gray}{(+4\%)}}
& 0.0525 & 0.0449 & 0.0213 & 0.0201 & 0.0631 & 0.0169 & 0.0365 \\
Transfer-Na\"{i}ve
& 0.0835 {\scriptsize\textcolor{gray}{(+4\%)}}
& 0.0799 {\scriptsize\textcolor{gray}{(+77\%)}}
& 0.0399 {\scriptsize\textcolor{gray}{(+51\%)}}
& 0.0283 {\scriptsize\textcolor{gray}{(+26\%)}}
& 0.0594 & 0.0476 & 0.0234 & 0.0217 & 0.0692 & 0.0183 & 0.0399 \\
\midrule
Transfer-Aware
& \textbf{0.0801}
& \textbf{0.0451}
& \textbf{0.0264}
& \textbf{0.0224}
& \textbf{0.0399}
& \textbf{0.0297}
& \textbf{0.0180}
& \textbf{0.0174}
& \textbf{0.0479}
& \textbf{0.0057}
& \textbf{0.0264} \\
\bottomrule
\end{tabular}}
\end{table}

\begin{table*}[t]
\centering
\caption{
Downstream performance across pretraining strategies. (*: p < 0.01, with Paired
Wilcoxon signed-rank test to Data-Proportional mixture pretraining baseline)}
\label{tab:downstream}
\vspace{4pt}
\resizebox{\textwidth}{!}{%
\begin{tabular}{l cc c c cc cc cc}
\toprule
& \multicolumn{3}{c}{\textbf{Classification (Accuracy (AUC) \%)}}
& & \multicolumn{6}{c}{\textbf{Segmentation (Dice \%)}} \\
\cmidrule(lr){2-4} \cmidrule(lr){6-11}
& \textbf{OASIS-1} & \textbf{SynBrain-PET} & \textbf{OrganMNIST3D}
& & \multicolumn{2}{c}{\textbf{AMOS}} & \multicolumn{2}{c}{\textbf{BraTS}} & \textbf{HECKTOR} & \textbf{BHSD} \\
\cmidrule(lr){2-2} \cmidrule(lr){3-3} \cmidrule(lr){4-4} \cmidrule(lr){6-7} \cmidrule(lr){8-9} \cmidrule(lr){10-10}  \cmidrule(lr){11-11}
\textbf{Pretraining}
  & Brain MRI & Brain PET & CT
  & & ABD-CT & ABD-MRI & Brain-T1 & Brain-T2 & HN-CT & HN-CT\\
\midrule
\multicolumn{10}{l}{\textit{No mixture}} \\
\quad From scratch
  & 65.1 (73.8) & 75.4 (95.8) & 84.6 (98.8)
  & & 76.0 & 62.2 & 45.2 & 57.8 & 62.4 & 53.4\\
\quad Single-Modality only
  & 67.4 (76.0) & 67.1 (92.8) & 88.6 (99.3)
  & & 79.3 & 68.3 & 51.3 & 62.1 & 66.7 & 55.6\\
\midrule
\multicolumn{10}{l}{\textit{Data mixture} ($T = 100\text{K}$)} \\
\quad Data-Proportional
  & 66.9 (73.6) & 69.7 (94.5) & 88.9 (99.4)
  & & 81.6 & 68.8 & 50.4 & 61.4 & 67.8 & 57.2\\
\quad \textbf{Transfer-Aware (Ours)}
  & \textbf{69.7 (78.4)*} & \textbf{78.1 (95.9)*} & \textbf{90.3 (99.4)*}
  & & \textbf{83.1}* & \textbf{69.4}* & \textbf{55.1}* & \textbf{65.4}* & \textbf{71.1}* & \textbf{61.2}\\
\bottomrule
\end{tabular}}
\end{table*}

\section{Experimental Design}
\label{sec:experimental_design}

We evaluate whether transfer-aware scaling laws select better 3D medical pretraining mixtures than standard allocation heuristics, and whether the resulting pretrained encoder transfers to downstream supervised tasks. Our study spans six heterogeneous 3D medical domains: abdominal CT (ABD-CT), abdominal MRI (ABD-MRI), brain MRI-T1 (BRAIN-T1), brain MRI-T2 (BRAIN-T2), head-and-neck CT (HN-CT), and head-and-neck PET (HN-PET). All volumes are resampled to $1.5\,\mathrm{mm}$ isotropic resolution and pretrained using a 3D masked autoencoder (MAE) proxy on $96^3$ patches. Full preprocessing, model architecture, and optimization details are provided in Appendix~\ref{app:experimental_details}.

\paragraph{Mixture strategies.}
We compare the proposed \textit{Transfer-Aware} allocation against three baselines that represent common data-mixture heuristics. \textit{Uniform} assigns equal mass to every domain, $h_i = 1/K$. \textit{Data-Proportional} assigns mass proportional to the number of available volumes in each domain, reflecting the common default when no scaling-law information is available. \textit{Transfer-Na\"ive} uses the fitted per-domain scaling laws but disables cross-domain transfer by setting $\tau_{ij}=\delta_{ij}$, isolating the contribution of the directed transfer matrix: it shares the same self-domain calibration as Transfer-Aware but cannot exploit source-to-target transfer. All strategies use the same total pretraining budget $T$, so differences in performance reflect allocation quality rather than compute.

\paragraph{Scaling-law estimation and optimization.}
Given these four strategies, we estimate the transfer-aware scaling law from $K^2=36$ proxy runs: $K=6$ self-domain runs and $K(K{-}1)=30$ directed source-to-target runs. Each configuration is evaluated at budgets $B \in \{10K, 50K, 100K, 200K\}$ volume draws, and validation MAE loss is measured on the target domain. We fit self-domain parameters $(E_i, C_i, \beta_i)$ and directed transfer coefficients $\tau_{ij}$ as described in Section~\ref{sec:fitting_and_optimization}, then optimize the mixture over the constrained simplex with floor $\epsilon{=}0.05$ to obtain $\bm{h}^*$. Unless otherwise stated, all target domains are weighted equally in the objective. To interpret the resulting allocation structure, we decompose each mixture into target-specific effective budgets $T_{\mathrm{eff},i} = \sum_j \tau_{ij} h_j T$, which measure how much direct and transferred pretraining signal each target receives (Section~\ref{sec:scaling_law_analysis}).

\paragraph{Downstream validation.}
Beyond the MAE proxy objective, we test whether the derived mixture structure transfers to supervised clinical tasks by finetuning the pretrained ViT encoders on both dense prediction and image-level recognition benchmarks. For segmentation, we evaluate on AMOS 2022 for 16-class abdominal organ segmentation (reporting CT and MRI subsets separately) \cite{ji2022amos}, BraTS 2023 Adult Glioma for multi-label brain tumor segmentation \cite{adewole2023brain}, HECKTOR 2025 and BHSD for head-and-neck CT lesion segmentation \cite{quetinautomatic,wu2023bhsd}. For classification, we use three benchmarks that probe complementary aspects of 3D representation transfer: OASIS-1 for structural brain MRI-T1 classification related to aging and cognitive impairment \cite{marcus2007open}, SynBrain-PET for brain PET classification testing transfer to functional/metabolic imaging, and OrganMNIST3D for 3D abdominal organ recognition from CT volumes \cite{yang2023medmnist}. Across all tasks, segmentation decoders, skip connections, and classification heads are randomly initialized, and all finetuning protocols are held fixed across pretraining strategies. We report Dice (\%) for segmentation and accuracy with AUC for classification.
 
\section{Scaling Law Analysis}
\label{sec:scaling_law_analysis}


\paragraph{The learned transfer structure is asymmetric.}
\begin{wrapfigure}{r}{0.4\textwidth}
\centering
\includegraphics[width=0.38\textwidth]{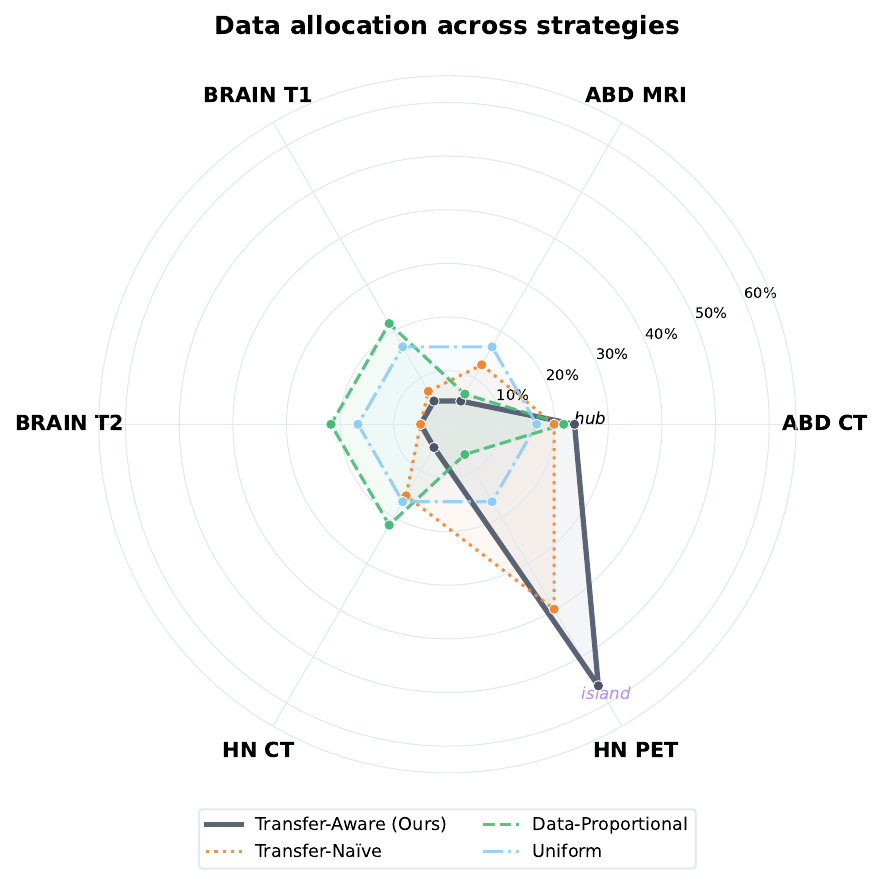}
\caption{
\textbf{Data allocation across strategies.}
Transfer-Aware concentrates budget on the hub (ABD-CT) and island (HN-PET), 
while heuristics spread budget uniformly or by dataset size.
}
\label{fig:allocation_radar}
\vspace{-20pt}
\end{wrapfigure} 
Figure~\ref{fig:observation}(c) shows the estimated directed transfer matrix $\tau_{ij}$ across the six 3D medical domains. Transfer is highly non-uniform and asymmetric: some sources provide broad benefit across targets, while others receive limited external support. In particular, ABD-CT acts as a high-yield source, contributing substantially to multiple targets, whereas HN-PET behaves as a low-transfer island and relies primarily on direct allocation. This structure is reflected in the effective-token decomposition in Table~\ref{tab:teff_decomposition}, and a non-uniform optimized mixture visualized in Figure~\ref{fig:allocation_radar}. At $T=100K$, ABD-CT receives $23.7\%$ of the direct budget but provides the largest cross-domain contribution to several targets, including ABD-MRI, BRAIN-T1, and BRAIN-T2. In contrast, HN-PET receives $56.3\%$ of the direct budget because its incoming transfer from other domains is weak. These results show that the optimized mixture is not simply proportional to dataset size or uniformly spread across domains; it reflects a learned source--target transfer structure.

\paragraph{Transfer-aware allocation improves MAE pretraining.}
Table~\ref{tab:pretraining_mae} compares Transfer-Aware allocation with Uniform, Data-Proportional, and Transfer-Na\"ive baselines under the same total budget. Transfer-Aware achieves the lowest mean MAE loss at every budget shown in Figure~\ref{fig:main_results}(a). The advantage is largest in the intermediate, data-bound regime: at $B=50K$, Transfer-Aware obtains a mean loss of $0.0451$, compared with $0.0671$ for Uniform, $0.0712$ for Data-Proportional, and $0.0799$ for Transfer-Na\"ive, corresponding to a $58\%$ relative improvement over Data-Proportional. At $B=100K$, Transfer-Aware reaches $0.0264$, improving over Uniform, Data-Proportional, and Transfer-Na\"ive by $31\%$, $38\%$, and $51\%$, respectively. The comparison with Transfer-Na\"ive is especially important because it uses the same per-domain scaling calibration but disables cross-domain transfer by setting $\tau_{ij}=\delta_{ij}$. The large gap between Transfer-Aware and Transfer-Na\"ive therefore isolates the benefit of modeling directed transfer, rather than only fitting separate per-domain scaling curves. The per-domain results at $B=100K$ further show that the gain is broad-based: Transfer-Aware achieves the lowest MAE loss on all six domains. The largest reduction occurs on HN-PET, where loss decreases from $0.0169$ under Data-Proportional to $0.0057$ under Transfer-Aware, consistent with the learned allocation assigning more direct budget to domains that receive little incoming transfer.

\paragraph{The scaling law extrapolates to a held-out budget.}
To test whether the fitted parameters remain stable beyond the calibration regime, we fit the law using budgets $B \in \{10K, 50K, 100K\}$ and predict performance at the held-out budget $B=200K$. Figure~\ref{fig:main_results}(b) shows that the predicted average loss closely matches the observed value: $0.0227$ predicted versus $0.0224$ observed. Figure~\ref{fig:main_results}(c) further shows strong per-domain agreement, with correlation $r=0.989$ across the six domains. This supports the use of small proxy runs to estimate mixture parameters for larger pretraining budgets. At $B=200K$, the relative advantage of Transfer-Aware narrows compared with the intermediate-budget regime. This suggests that the proxy model may be approaching either the irreducible loss floor or a capacity-limited regime, where additional data provides diminishing returns regardless of allocation. We therefore interpret the strongest gains as evidence for transfer-aware allocation in the data-bound regime, while larger-model validation remains an important direction for future work.
 
\section{Downstream Finetuning Results}
\label{sec:downstream_results}

\paragraph{Transfer-aware pretraining improves classification.}
On image-level classification, Transfer-Aware achieves the best performance across all three benchmarks. On OASIS-1 brain MRI, it improves accuracy from $66.9\%$ under Data-Proportional to $69.7\%$, and AUC from $73.6\%$ to $78.4\%$. On SynBrain-PET, the gain is larger: Transfer-Aware reaches $78.1\%$ accuracy and $95.9\%$ AUC, compared with $69.7\%$ accuracy and $94.5\%$ AUC for Data-Proportional. This result is consistent with the scaling-law analysis, where PET behaves as a weak-transfer domain that benefits from targeted direct allocation. On OrganMNIST3D, Transfer-Aware also obtains the highest accuracy, improving from $88.9\%$ to $90.3\%$, while matching the strongest AUC at $99.4\%$. These results indicate that the optimized mixture improves global volume-level representations, not only voxel-level reconstruction.

\paragraph{Transfer-aware pretraining improves segmentation.}
The same trend holds for dense prediction. On AMOS, Transfer-Aware achieves $83.1$ Dice on abdominal CT and $69.4$ Dice on abdominal MRI, outperforming Data-Proportional by $1.5$ and $0.6$ points, respectively. The gains are larger on brain and head-and-neck tasks. On BraTS, Transfer-Aware improves over Data-Proportional from $50.4$ to $55.1$ Dice on Brain T1 and from $61.4$ to $65.4$ Dice on Brain T2. On head-and-neck CT segmentation, Transfer-Aware improves HECKTOR from $67.8$ to $71.1$ Dice and BHSD from $57.2$ to $61.2$ Dice. Across all six segmentation tasks, Transfer-Aware yields the best performance among the compared pretraining strategies.

\paragraph{Proxy improvements translate to downstream transfer.}
Overall, the downstream results mirror the proxy-loss ordering observed in Table~\ref{tab:pretraining_mae}: Transfer-Aware consistently outperforms Data-Proportional, which in turn generally improves over training from scratch. This agreement suggests that the MAE proxy is not merely optimizing reconstruction loss in isolation, but provides a useful signal for selecting pretraining mixtures that transfer to supervised classification and segmentation. The gains across MRI, CT, and PET tasks further indicate that modeling directed transfer is beneficial for learning broadly reusable 3D medical representations under a fixed pretraining budget.

\section{Discussion}
\label{sec:discussion}

Our transfer-aware scaling law demonstrates that heterogeneous 3D medical data mixtures can be optimized analytically rather than searched exhaustively. The framework achieves the lowest MAE loss at every budget tested, extrapolates to unseen budgets with $1.2\%$ error ($r{=}0.989$), and transfers to downstream clinical tasks. The learned allocations reveal interpretable hub-and-island structure that provides actionable guidance for practitioners training unified 3D medical foundation models.

\textbf{Limitations.} Two modeling assumptions merit further examination. First, our mixture law assumes transfer contributions compose linearly ($T_{\mathrm{eff},i} = \sum_j \tau_{ij} h_j T$), which cannot capture synergistic or interfering interactions between source domains. The strong empirical performance suggests first-order transfer dominates in our six-domain setting, but nonlinear interaction terms may become important as domain count or modality diversity increases. Second, the transfer matrix $\bm{\tau}$ is estimated once from small-scale proxy runs and held constant across budgets. Our extrapolation results validate this within the $10K$--$200K$ range, but transfer structure could evolve at substantially larger scales as models shift from learning broadly transferable low-level features to more domain-specific representations. Modeling $\tau_{ij}(T)$ as a budget-dependent function is a natural extension for future work. A broader impact discussion is provided in Appendix~\ref{app:broader_impact}.

\section{Conclusion}
\label{sec:conclusion}

We introduced a transfer-aware scaling law for 3D medical data mixture optimization. By modeling both domain-specific scaling behavior and directed cross-domain transfer, our framework converts heterogeneous pretraining mixture design from a heuristic choice into an explicit compute-aware optimization problem. Across six diverse CT, MRI, and PET domains, the learned allocation improves MAE pretraining loss, extrapolates accurately to a held-out budget, and yields pretrained encoders that transfer better to downstream classification and segmentation tasks. These results highlight data allocation as a key design axis for scalable 3D medical foundation models and suggest that proxy-scale scaling laws can guide more efficient use of heterogeneous medical imaging data.

\bibliographystyle{splncs04}
\bibliography{neurips_2026}


\newpage
\appendix

\section{Experimental Details}
\label{app:experimental_details}

\subsection{Dataset Description}
\label{app:datasets}

\paragraph{Pretraining datasets.}
Our six pretraining domains are drawn from three publicly available 3D medical imaging collections spanning CT, MRI, and PET (Table~\ref{tab:pretrain_datasets}). To simulate realistic data imbalance across domains, we subsample each source dataset to a controlled size: 500 volumes for high-availability domains (ABD-CT, BRAIN-T1, BRAIN-T2, HN-CT) and 150 volumes for low-availability domains (ABD-MRI, HN-PET). This $3.3\times$ imbalance ratio reflects practical clinical settings where some modalities (e.g., CT) are far more abundant than others (e.g., PET or MRI), and ensures that data-proportional allocation produces meaningfully different mixtures from uniform allocation. The specific volume counts are chosen to be large enough for stable MAE training at our proxy scale with time and cost efficiency (i.e. Finished pretraining within a day for empirical observation) while remaining computationally tractable for $K^2 = 36$ proxy runs across four budget levels. All pretraining volumes are unlabeled and used exclusively for self-supervised MAE reconstruction.

\begin{table}[h]
\centering
\caption{Pretraining domain statistics. ``Available'' refers to the total unlabeled volumes in the source dataset; ``Selected'' is the number subsampled for our experiments. All volumes are resampled to $1.5\,\mathrm{mm}$ isotropic resolution.}
\label{tab:pretrain_datasets}
\begin{tabular}{llccl}
\toprule
\textbf{Domain} & \textbf{Modality} & \textbf{Available} & \textbf{Selected} & \textbf{Source} \\
\midrule
ABD-CT   & CT       & 1100 & 500 & AMOS 2022~\cite{ji2022amos} \\
ABD-MRI  & MRI      & 700 & 150 & AMOS 2022~\cite{ji2022amos} \\
BRAIN-T1 & MRI (T1 + ce) & 2052  & 500 & BraTS 2023~\cite{adewole2023brain} \\
BRAIN-T2 & MRI (T2 + FLAIR) & 2052  & 500 & BraTS 2023~\cite{adewole2023brain} \\
HN-CT    & CT       &  700  & 500 & HECKTOR 2025~\cite{quetinautomatic} \\
HN-PET   & PET      &  700  & 150 & HECKTOR 2025~\cite{quetinautomatic} \\
\bottomrule
\end{tabular}
\end{table}

\paragraph{Downstream finetuning datasets.}
To validate that the MAE-derived mixture structure transfers beyond the proxy objective, we finetune pretrained encoders on six supervised downstream tasks spanning segmentation and classification across all three modalities (Table~\ref{tab:downstream_datasets}). Downstream subjects are held out from pretraining to avoid data leakage. For segmentation, we use the labeled subsets of the same source collections (AMOS, BraTS, HECKTOR) as well as an additional stroke CT dataset (CPAISD). For classification, we use three benchmarks that probe complementary aspects of 3D representation transfer: structural brain MRI (OASIS-1), functional brain PET (SynBrain-PET), and abdominal CT organ recognition (OrganMNIST3D).

\begin{table}[h]
\centering
\caption{Downstream finetuning datasets. Segmentation tasks report 
Dice (\%); classification tasks report accuracy and AUC. All 
finetuning protocols are held fixed across pretraining strategies.}
\label{tab:downstream_datasets}
\resizebox{\columnwidth}{!}{%
\begin{tabular}{lccccl}
\toprule
\textbf{Dataset} & \textbf{Task} & \textbf{Classes} & \textbf{Number of Subjects (Train/Val/Test)} & \textbf{Metric} & \textbf{Domain} \\
\midrule
\multicolumn{6}{l}{\textit{Segmentation}} \\
\quad AMOS 2022~\cite{ji2022amos}     & Organ seg.  & 16             & 200 (160/20/20)   & Dice & ABD-CT / MRI \\
\quad BraTS 2023~\cite{adewole2023brain}  & Tumor seg.  & 3 & 480 (400/40/40)   & Dice & BRAIN-T1 / T2 \\
\quad HECKTOR 2025~\cite{quetinautomatic} & Tumor seg. & 3 & 524 (418/53/53) & Dice & HN-CT \\
\quad BHSD 2024~\cite{wu2023bhsd}     & Lesion seg. & 3              &  92 (72/10/10)  & Dice & HN-CT \\
\midrule
\multicolumn{6}{l}{\textit{Classification}} \\
\quad OASIS-1~\cite{marcus2007open}  & Brain class. & 2             & 235 (165/35/35)  & Acc / AUC & BRAIN-T1 \\
\quad SynBrain-PET~\cite{barnard2025fdg} & PET class. & 5            & 500 (348/76/76)    & Acc / AUC & HN-PET \\
\quad OrganMNIST3D~\cite{yang2023medmnist} & Organ class. & 11      & 1,743 (972/161/610) & Acc / AUC & ABD-CT \\
\bottomrule
\end{tabular}
}
\end{table}

\subsection{Preprocessing}
\label{app:preprocessing}

\paragraph{Pretraining preprocessing.}
All pretraining volumes are reoriented to RAS and resampled to $1.5\,\mathrm{mm}$ isotropic resolution using trilinear interpolation. Domain-specific intensity normalization maps all volumes to $[-1, 1]$ based on modality and anatomy, as summarized in Table~\ref{tab:preprocessing}. After normalization, foreground regions are cropped, volumes are padded to at least $96^3$ voxels, and a random $96^3$ patch is extracted for training. Light augmentation is applied during pretraining: random intensity scaling ($\pm 10\%$, $p=0.3$), random contrast adjustment ($p=0.2$), and random Gaussian noise ($\sigma=0.02$, $p=0.15$). A $75\%$ random patch masking ratio is applied for masked autoencoding.

\begin{table}[h]
\centering
\caption{Domain-specific intensity preprocessing. All domains share the same spatial preprocessing (RAS orientation, $1.5\,\mathrm{mm}$ isotropic resampling, foreground cropping, $96^3$ patching).}
\label{tab:preprocessing}
\resizebox{0.9\columnwidth}{!}{%
\begin{tabular}{lccc}
\toprule
\textbf{Domain} & \textbf{Family} & \textbf{Intensity normalization} & \textbf{Output range} \\
\midrule
ABD-CT   & CT  & HU window $[-160, 240]$, linear rescale & $[-1, 1]$ \\
HN-CT    & CT  & HU window $[-200, 300]$, linear rescale & $[-1, 1]$ \\
ABD-MRI  & MRI & $z$-norm (nonzero), clip $\pm 3\sigma$, rescale & $[-1, 1]$ \\
BRAIN-T1 & MRI & $z$-norm (nonzero), clip $\pm 5\sigma$, rescale & $[-1, 1]$ \\
BRAIN-T2 & MRI & $z$-norm (nonzero), clip $\pm 5\sigma$, rescale & $[-1, 1]$ \\
HN-PET   & PET & $z$-norm (nonzero), clip $\pm 5\sigma$, rescale & $[-1, 1]$ \\
\bottomrule
\end{tabular}
}
\end{table}

\paragraph{Finetuning preprocessing.}
Finetuning transforms are designed to match the pretraining intensity and resolution pipeline exactly, ensuring that pretrained encoder weights see the same input distribution. Labels are resampled using nearest-neighbor interpolation to preserve discrete class values. Table~\ref{tab:finetune_preprocessing} summarizes the domain-specific finetuning configuration. The key design choices are:
\begin{itemize}
    \item \textbf{BraTS (BRAIN-T1/T2):} Labels $\{0, 1, 2, 4\}$ are converted to three overlapping binary channels (WT, TC, ET) \emph{after} spatial transforms and random cropping, enabling multi-label sigmoid training. The conversion is placed after \texttt{RandCropByPosNegLabeld} in MONAI because the cropper requires single-channel integer labels to identify foreground voxels.
    \item \textbf{HN-CT (HECKTOR, BHSD):} Small tumor/lesion targets use \texttt{pos=4, neg=1} with \texttt{num\_samples=4} to ensure sufficient foreground exposure during training.
    \item \textbf{AMOS (ABD-CT/MRI):} Standard organ segmentation with \texttt{pos=1, neg=1} and \texttt{num\_samples=2}.
\end{itemize}

\begin{table}[h]
\centering
\caption{Finetuning configuration per downstream domain.}
\label{tab:finetune_preprocessing}
\resizebox{0.9\columnwidth}{!}{%
\begin{tabular}{lcccccc}
\toprule
\textbf{Domain} & \textbf{Intensity} & \textbf{pos} & \textbf{neg} & \textbf{samples} & \textbf{Loss} & \textbf{Dice metric} \\
\midrule
ABD-CT   & HU $[-160, 240] \to [-1, 1]$       & 1 & 1 & 2 & DiceCE (softmax) & incl.\ background \\
ABD-MRI  & $z$-norm, $\pm 3\sigma \to [-1, 1]$ & 1 & 1 & 2 & DiceCE (softmax) & incl.\ background \\
BRAIN-T1 & $z$-norm, $\pm 5\sigma \to [-1, 1]$ & 4 & 1 & 2 & DiceCE (sigmoid) & excl.\ background \\
BRAIN-T2 & $z$-norm, $\pm 5\sigma \to [-1, 1]$ & 4 & 1 & 2 & DiceCE (sigmoid) & excl.\ background \\
HN-CT    & HU $[-200, 300] \to [-1, 1]$        & 4 & 1 & 4 & DiceCE (softmax) & excl.\ background \\
\bottomrule
\end{tabular}
}
\end{table}

\paragraph{Finetuning augmentation.}
All segmentation tasks share the same augmentation pipeline: random intensity shift ($\pm 0.1$, $p=0.5$) and random affine transform (rotation $\pm \pi/30$, scale $\pm 10\%$, $p=1.0$). No augmentation is applied during validation.

\paragraph{Inference.}
Training-time validation uses sliding-window inference with \texttt{sw\_batch\_size}$=1$ and overlap $=0.25$ for computational efficiency. Final evaluation uses \texttt{sw\_batch\_size}$=4$ and overlap $=0.7$ for higher accuracy. All predictions and metrics are computed at the resampled $1.5\,\mathrm{mm}$ resolution.

\subsection{Model Architecture}
\label{app:architecture}

The MAE proxy model uses a Vision Transformer (ViT)~\cite{dosovitskiy2020image} with $768$-dimensional hidden size, $12$ transformer layers, $12$ attention heads, MLP dimension $3072$, and $16^3$ patch size. The ViT encoder is embedded within the UNETR~\cite{hatamizadeh2022unetr} framework, which provides a CNN decoder for masked patch reconstruction during pretraining. Each domain uses a dedicated convolutional input 
adapter that projects single-channel $96^3$ volumes to the shared patch embedding space. During pretraining, $75\%$ of patches are randomly masked and the model reconstructs masked patches using per-patch MSE loss \cite{he2022masked}.

\begin{table}[h]
\centering
\caption{UNETR parameter breakdown. Only the ViT encoder is initialized from the pretrained MAE checkpoint; the CNN decoder and task-specific heads are randomly initialized for each downstream task.}
\label{tab:param_split}
\begin{tabular}{lrc}
\toprule
\textbf{Component} & \textbf{Parameters} & \textbf{Initialization} \\
\midrule
ViT encoder (\texttt{vit.*}) & 116,679,936 & Pretrained (MAE) \\
CNN decoder (\texttt{encoder1--4, decoder2--5}) & 4,443,140 & Random \\
Output head (\texttt{out}) & varies by task & Random \\
\midrule
\textbf{Total} & $\sim$121M & -- \\
\bottomrule
\end{tabular}
\end{table}

\subsection{Pretraining Optimization}
\label{app:pretrain_optim}

All pretraining runs use AdamW~\cite{loshchilov2017decoupled} with learning rate $1.5 \times 10^{-4}$, batch size $4$, and $50$ epochs. Training samples are drawn according to the prescribed mixture weights $\bm{h}$ using weighted random sampling without replacement within each epoch. We use a cache rate of $1.0$ for MONAI CacheDataset to maximize throughput. Each configuration is run with $2$--$3$ random seeds and results are averaged to reduce variance in both scaling-law fitting and final evaluation.

\subsection{Downstream Model Initialization}
\label{app:model_init}

\paragraph{Segmentation: UNETR with pretrained ViT encoder.}
For segmentation tasks, we use UNETR~\cite{hatamizadeh2022unetr}, which combines a ViT encoder with a CNN decoder connected through skip connections at multiple resolutions. The pretrained MAE checkpoint stores the full UNETR state dict (ViT encoder + CNN reconstruction decoder). During finetuning initialization, we load \textbf{only the ViT encoder weights} (\texttt{vit.*} keys) and \textbf{randomly initialize} the CNN decoder branches (\texttt{encoder1}--\texttt{encoder4}, \texttt{decoder2}--\texttt{decoder5}, \texttt{out}). This is deliberate: the ViT encoder learns spatial representations from MAE that transfer well to segmentation, but the CNN decoder was trained for pixel-level reconstruction (regression), which differs fundamentally from segmentation (per-voxel classification). Loading reconstruction decoder weights can cause negative transfer. The segmentation output head is initialized with the correct number of output channels for the target task. Weight loading proceeds as follows:
\begin{enumerate}
    \item Load the MAE checkpoint containing \texttt{unetr\_state\_dict} and metadata (\texttt{scale\_factor}, \texttt{num\_layers}).
    \item Build a fresh UNETR with \texttt{out\_channels = num\_classes} for the target task.
    \item Extract all keys matching \texttt{vit.*} from the checkpoint and load them into the new UNETR's ViT encoder, verifying shape compatibility.
    \item All remaining parameters (CNN decoder, skip connections, output head) retain random initialization.
\end{enumerate}

\paragraph{Classification: ViT encoder with MLP head.}
For classification tasks, we extract the same pretrained ViT encoder and attach a lightweight MLP classification head. The ViT processes the input $96^3$ volume into a sequence of patch embeddings, which are globally average-pooled to produce a single feature vector of dimension $d = 768$. The pooled feature is passed through LayerNorm and an MLP head to map it to the target classes:
\begin{equation*}
    \hat{y} = f_{\mathrm{MLP}}\left(\mathrm{LN}\left(\mathrm{GAP}(\mathrm{ViT}(x))\right)\right),
\end{equation*}
where GAP denotes global average pooling over the patch token sequence. The MLP head is randomly initialized, while the ViT encoder is loaded from the same pretrained checkpoint and finetuned according to the downstream training schedule.

\paragraph{From-scratch baseline.}
Both segmentation and classification baselines use the identical architecture with all parameters randomly initialized (no pretrained weights), isolating the effect of pretraining from architectural differences.

\subsection{Downstream Finetuning Protocol}
\label{app:finetune_protocol}

Finetuning uses AdamW with learning rate $1e{-4}$, batch size $2$, and $200$ epochs for all downstream tasks. For segmentation, we use DiceCE loss with softmax for multi-class tasks (AMOS, HECKTOR, BHSD) and DiceCE loss with sigmoid for multi-label segmentation (BraTS). For classification, we use cross-entropy loss. All finetuning protocols (i.e. optimizer, learning rate, epochs, augmentation, and evaluation) are held fixed across pretraining strategies to ensure fair comparison.

\paragraph{Segmentation augmentation.}
Training augmentation follows~\cite{lee20223d}: foreground cropping, spatial padding to $96^3$, random cropping via \texttt{RandCropByPosNegLabeld} (pos$=1$, neg$=1$ for organ segmentation; pos$=4$, neg$=1$ for small lesion/tumor tasks) in MONAI package, random intensity shifts ($\pm 0.1$, $p=0.5$), and random affine transforms (rotation $\pm \pi/30$, scale $\pm 10\%$). Validation uses sliding-window inference with overlap $0.5$ for both validation and final evaluation.

\subsection{Fixed Model Size}
\label{app:fixed_model_size}

We fix model size and optimize over data mixture and budget, rather than jointly optimizing model size as in Chinchilla-style scaling laws~\cite{hoffmann2022training}. This is deliberate: 3D medical imaging at $96^3$ resolution is GPU-memory-bound, and model architecture is typically standardized across benchmarks~\cite{hatamizadeh2022unetr,lee20223d}. The practical question facing practitioners is how to allocate a fixed data budget across heterogeneous domains, not how to trade off model size against data. Our scaling law therefore models $L(\bm{h}, T)$ at fixed model size $N$, making the surrogate directly actionable for the regime where most 3D medical pretraining operates. Extending the framework to jointly optimize $L(N, \bm{h}, T)$ is a natural direction for future work as hardware constraints relax.

\section{Additional Results}
\label{app:additional_results}

\subsection{Domain Characterization}
\label{app:domain_characterization}

\begin{table}[h]
\centering
\caption{Domain characterization from the transfer matrix and prescribed allocations.}
\label{tab:domain_roles}
\begin{tabular}{lccccc}
\toprule
\textbf{Domain} & $\bar{\tau}_{\text{out}}$ & $\bar{\tau}_{\text{in}}$ & \textbf{Role} & $h^*_{\text{TA}}$ & $h_{\text{DP}}$ \\
\midrule
ABD-CT   & 0.29 & 0.32 & Hub    & 23.7\% & 21.7\% \\
ABD-MRI  & 0.24 & 0.22 & --     &  5.0\% &  6.5\% \\
BRAIN-T1 & 0.21 & 0.25 & Easy   &  5.0\% & 21.7\% \\
BRAIN-T2 & 0.26 & 0.21 & Easy   &  5.0\% & 21.7\% \\
HN-CT  & 0.25 & 0.25 & --     &  5.0\% & 21.7\% \\
HN-PET & 0.09 & 0.09 & Island & 56.3\% &  6.5\% \\
\bottomrule
\end{tabular}
\end{table}

Table~\ref{tab:domain_roles} summarizes the role of each domain as determined by the estimated transfer matrix. ABD-CT acts as a hub with the highest mean outgoing transfer ($\bar{\tau}_{\text{out}} = 0.29$), contributing meaningfully to four other domains. HN-PET is an island with negligible incoming and outgoing transfer ($\bar{\tau} = 0.09$), reflecting its low-resolution metabolic signal that shares little representational structure with the other five domains. BRAIN-T1 and BRAIN-T2 are classified as ``easy'' domains due to their rapid saturation ($\beta \in [0.18, 0.24]$), indicating strong anatomical regularity that is captured with relatively little data.

\subsection{Transfer Amplification}
\label{app:amplification}

\begin{table}[h]
\centering
\caption{Transfer amplification at $B\!=\!100K$. Floor domains ($h\!=\!5\%$) achieve $3.7$--$6.1\times$ effective budget through cross-domain transfer, and still outperform Data-Proportional.}
\label{tab:amplification}
\begin{tabular}{lrrrccc}
\toprule
\textbf{Domain} & $h_i \!\cdot\! T$ & $T_{\text{eff}}$ & Boost & TA (Ours) & DP & $\Delta$ \\
\midrule
ABD-CT   & 23,700 & 33,652 & 1.4$\times$ & \textbf{0.399} & 0.525 & $-$24\% \\
ABD-MRI  &  5,000 & 20,928 & 4.2$\times$ & \textbf{0.297} & 0.449 & $-$34\% \\
BRAIN-T1 &  5,000 & 20,212 & 4.0$\times$ & \textbf{0.180} & 0.213 & $-$15\% \\
BRAIN-T2 &  5,000 & 18,334 & 3.7$\times$ & \textbf{0.174} & 0.201 & $-$13\% \\
HN-CT  &  5,000 & 30,289 & 6.1$\times$ & \textbf{0.479} & 0.631 & $-$24\% \\
HN-PET & 56,300 & 59,248 & 1.1$\times$ & \textbf{0.057} & 0.169 & $-$66\% \\
\bottomrule
\end{tabular}
\end{table}

Table~\ref{tab:amplification} shows the transfer amplification effect at $B = 100K$. Floor domains ($h = 5\%$) receive only $5{,}000$ direct volume draws but achieve $3.7$--$6.1\times$ effective budget through cross-domain transfer. Despite this minimal direct allocation, all floor domains achieve lower MAE loss under Transfer-Aware than under Data-Proportional, demonstrating a Pareto improvement across all domains simultaneously. The largest improvement is on HN-PET ($-66\%$), where Transfer-Aware allocates $56.3\%$ of the budget versus only $6.5\%$ under Data-Proportional.

\subsection{Scaling Exponent Comparison}
\label{app:scaling_exponents}

\begin{table}[h]
\centering
\caption{Fitted scaling exponents ($L_i = C_i / T^{\beta_i}$). Transfer-Aware achieves $1.31\times$ higher mean $\beta$.}
\label{tab:scaling_params}
\begin{tabular}{lcc|cc|c}
\toprule
& \multicolumn{2}{c|}{\textbf{TA (Ours)}} & \multicolumn{2}{c|}{\textbf{DP}} & \\
\textbf{Domain} & $\beta$ & $C$ & $\beta$ & $C$ & Ratio \\
\midrule
ABD-CT   & 0.407 &  5.02 & 0.302 & 2.05 & 1.35$\times$ \\
ABD-MRI  & 0.487 &  9.51 & 0.352 & 3.01 & 1.38$\times$ \\
BRAIN-T1 & 0.209 &  0.21 & 0.173 & 0.15 & 1.21$\times$ \\
BRAIN-T2 & 0.243 &  0.30 & 0.225 & 0.27 & 1.08$\times$ \\
HN-CT  & 0.436 &  8.12 & 0.330 & 3.24 & 1.32$\times$ \\
HN-PET & 0.724 & 39.80 & 0.466 & 4.96 & 1.55$\times$ \\
\midrule
\textbf{Mean} & \textbf{0.425} & \textbf{4.06} & 0.325 & 1.77 & 1.31$\times$ \\
\bottomrule
\end{tabular}
\end{table}

Table~\ref{tab:scaling_params} compares the fitted scaling exponents under Transfer-Aware and Data-Proportional allocation. Transfer-Aware achieves a $1.31\times$ higher mean scaling exponent ($\beta_{\text{TA}} = 0.425$ vs.\ $\beta_{\text{DP}} = 0.325$), indicating that each additional unit of compute yields more improvement under Transfer-Aware allocation. The largest exponent ratio is on HN-PET ($1.55\times$), reflecting that direct investment in the island domain produces the steepest returns. Brain domains show the smallest ratios ($1.08$--$1.21\times$), consistent with their rapid saturation.

\subsection{Effective Budget Predicts Loss}
\label{app:effective_budget}

\begin{table}[h]
\centering
\caption{Effective budget ratio predicts loss ratio with Pearson $r\!=\!0.992$ at $B\!=\!100K$.}
\label{tab:effective_budget}
\begin{tabular}{lccc}
\toprule
\textbf{Domain} & $T_{\text{eff}}^{\text{TA}} / T_{\text{eff}}^{\text{DP}}$ & $L^{\text{DP}} / L^{\text{TA}}$ & $\Delta$ \\
\midrule
ABD-CT   & 0.72 & 1.32 & $-$24\% \\
ABD-MRI  & 0.69 & 1.51 & $-$34\% \\
BRAIN-T1 & 0.45 & 1.18 & $-$15\% \\
BRAIN-T2 & 0.46 & 1.16 & $-$13\% \\
HN-CT  & 0.71 & 1.32 & $-$24\% \\
HN-PET & 3.80 & 2.96 & $-$66\% \\
\bottomrule
\end{tabular}
\end{table}

Table~\ref{tab:effective_budget} validates the transfer matrix by comparing the effective budget ratio between Transfer-Aware and Data-Proportional allocations with the corresponding loss ratio at $B = 100K$. The two quantities correlate at $r = 0.992$, confirming that $\tau_{ij}$, estimated independently from pure-domain proxy runs, correctly predicts which domains gain effective compute and by how much. HN-PET shows the most extreme effective budget ratio ($3.80\times$) and the largest loss improvement ($2.96\times$).

\subsection{Per-Domain Scaling Curves}
\label{app:per_domain_scaling}

\begin{figure*}[h]
\centering
\includegraphics[width=\textwidth]{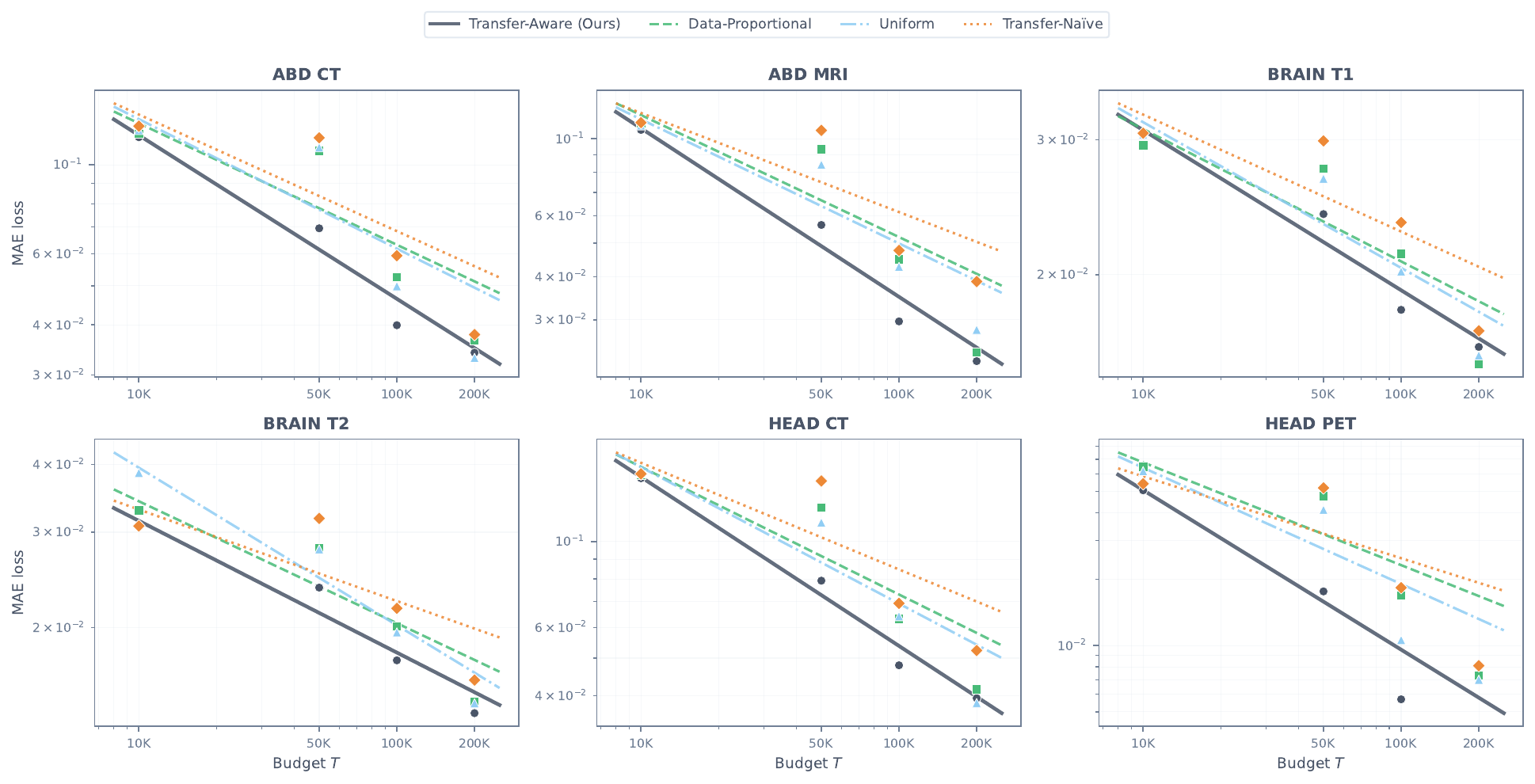}
\caption{Per-domain scaling laws (log-log) comparing transfer-aware and data-proportional allocation. Transfer-aware achieves a steeper scaling exponent $\beta$ on all 6 domains, with the largest gains on HEAD~PET ($\beta\!: 0.724$ vs.\ $0.466$) and ABD~MRI ($\beta\!: 0.487$ vs.\ $0.352$). BRAIN~T1 and BRAIN~T2 show the smallest differences, consistent with these domains being near saturation at low $\beta$ values. The $3.5\times$ heterogeneity in $\beta$ across domains ($0.209$--$0.724$) confirms that a single shared scaling exponent, as assumed by prior work, is insufficient for heterogeneous medical data mixtures.}
\label{fig:per_domain_scaling}
\end{figure*}

Figure~\ref{fig:per_domain_scaling} shows per-domain MAE loss across all four mixture strategies on log-log axes. Transfer-Aware achieves a steeper scaling exponent on all six domains, with the largest gains on HN-PET ($\beta: 0.724$ vs.\ $0.466$) and ABD-MRI ($\beta: 0.487$ vs.\ $0.352$). Brain domains show the smallest inter-strategy gap, consistent with their rapid saturation at low $\beta$ values. The $3.5\times$ heterogeneity in $\beta$ across domains ($0.209$--$0.724$) confirms that a single shared scaling exponent, as assumed by prior work, is insufficient for heterogeneous 3D medical data mixtures.

\subsection{Floor Constraint Ablation}
\label{app:hfloor}

\begin{figure*}[h]
\centering
\includegraphics[width=\textwidth]{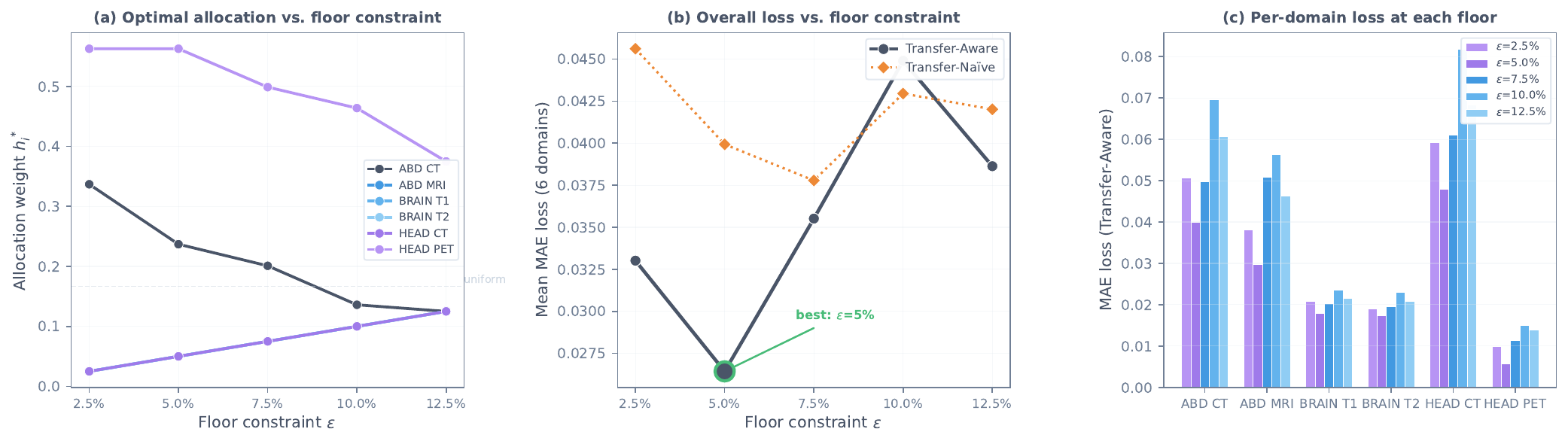}
\caption{
\textbf{Floor constraint ablation.}
\textbf{(a)} As the floor $\epsilon$ increases, the optimizer loses flexibility: HEAD PET allocation drops from $56.3\%$ to $37.5\%$ and ABD CT from $33.7\%$ to $12.5\%$.
\textbf{(b)} Mean MAE loss is minimized at $\epsilon{=}5\%$; higher floors force budget into domains that receive sufficient signal through transfer, degrading overall performance.
\textbf{(c)} Per-domain losses confirm that $\epsilon{=}5\%$ achieves the best or near-best loss on every domain simultaneously.
}
\label{fig:hfloor}
\end{figure*}

Figure~\ref{fig:hfloor} shows the effect of varying the floor constraint $\epsilon$ from $2.5\%$ to $12.5\%$. The optimal value is $\epsilon = 5\%$: lower floors ($2.5\%$) over-concentrate budget on just two domains (ABD-CT and HN-PET), while higher floors ($\ge 10\%$) force budget into domains that receive sufficient signal through transfer, eliminating the advantage of the transfer matrix. At $\epsilon = 10\%$, Transfer-Aware actually underperforms Transfer-Na\"ive, confirming that an over-constrained floor removes the optimizer's ability to exploit transfer structure.

\section{Compute Cost}
\label{app:compute}

The total estimation cost consists of $K^2 = 36$ pure-domain proxy runs at four budgets each. Each proxy run trains a $12$M-parameter ViT for $50$ epochs, requiring approximately $10-20$ GPU-hours on a single NVIDIA A100 depending on domain size and resolution. Once the scaling law is fitted, optimizing a new mixture for any budget takes less than one second on a single CPU core (100 SLSQP restarts). By comparison, a single full-scale pretraining run at $B = 200K$ requires approximately 3 days, and an exhaustive grid search over even a coarse $5$-point discretization of the $5$-dimensional mixture simplex would require $5^5 = 3{,}125$ such runs.

\section{Broader Impact}
\label{app:broader_impact}

By replacing exhaustive mixture search with principled scaling-law optimization, our framework reduces the compute cost of training 3D medical foundation models, lowering the barrier for resource-constrained institutions. However, the framework optimizes a self-supervised proxy (MAE loss) rather than clinical outcomes directly. Practitioners should validate downstream performance on their specific clinical tasks before deployment. Additionally, the optimized mixture inherits any biases present in the source datasets (i.e If certain patient populations are underrepresented in specific imaging domains), the allocation will not address this imbalance. Incorporating fairness-aware objectives into the mixture optimization is an important direction for future work.



\end{document}